\documentclass[letterpaper, 10 pt, conference]{ieeeconf}  

\IEEEoverridecommandlockouts                              

\usepackage{amsmath} 
\usepackage{amssymb}  
\usepackage{graphicx} 
\usepackage{booktabs}
\usepackage{balance}
\usepackage{hyperref}

\title{\LARGE \bf 
DB-TSDF: Directional Bitmask-based Truncated Signed Distance Fields for Efficient Volumetric Mapping 
}

\author{José E. Maese$^{1}$, Luis Merino$^{1}$ and Fernando Caballero$^{1}$
\thanks{*This work was supported by the grants PICRA 4.0 (PLEC2023-010353), funded by the Spanish Ministry of Science and Innovation and the Spanish Research Agency (MCIN/AEI/10.13039/501100011033); and COBUILD (PID2024-161069OB-C31), funded by the  panish Ministry of Science, Innovation and Universities, the Spanish Research Agency (MICIU/AEI/10.13039/501100011033) and the European Regional Development Fund (FEDER, UE).}
\thanks{$^{1}$The authors are with the Service Robotics Laboratory, Universidad Pablo de Olavide, Seville, Spain. {\tt\small \{jemaealv,lmercab,fcaballero\}@upo.es}}%
}

\begin{document}

\maketitle
\thispagestyle{empty}
\pagestyle{empty}

\begin{abstract}

This paper presents a high-efficiency, CPU-only volumetric mapping framework based on a Truncated Signed Distance Field (TSDF). The system incrementally fuses raw LiDAR point-cloud data into a voxel grid using a directional bitmask-based integration scheme, producing dense and consistent TSDF representations suitable for real-time 3D reconstruction. A key feature of the approach is that the processing time per point-cloud remains constant, regardless of the voxel grid resolution, enabling high resolution mapping without sacrificing runtime performance. In contrast to most recent TSDF/ESDF methods that rely on GPU acceleration, our method operates entirely on CPU, achieving competitive results in speed. Experiments on real-world open datasets demonstrate that the generated maps attain accuracy on par with contemporary mapping techniques. The source code is publicly available at \url{https://github.com/robotics-upo/DB-TSDF}

\end{abstract}

\section{Introduction}
   
Volumetric mapping is a fundamental capability in mobile robotics, supporting tasks such as collision avoidance, motion planning, and the construction of consistent world models under real-time constraints. Point clouds and occupancy grids remain widely used on CPU-only platforms, as their simple data structures allow efficient processing without specialized hardware. However, they are prone to aliasing at high resolutions and often produce geometric artifacts that hinder downstream processing. Truncated Signed Distance Fields (TSDFs) address these limitations by storing per-voxel distances to the nearest surface and providing smooth proximity information. Despite their advantages, many existing TSDF and ESDF (Euclidean Signed Distance Fields) pipelines rely heavily on GPU acceleration or exhibit computational costs on the CPU that grow unfavorably with map resolution and update rate.

This work introduces DB-TSDF, a mapping method that integrates TSDFs using a directional bitmask representation specifically designed for fast operation on a discrete voxel grid using only the CPU. Each voxel encodes a compact 32-bit distance mask, a sign flag, and a hit counter. For each LiDAR return, the system selects a precomputed, direction-dependent kernel applied over a fixed neighborhood. A single bitwise AND operation per voxel updates the mask, while a directional shadow mechanism assigns occupied or free-space evidence. Since the kernel size remains constant, the integration time per scan is bounded and remains largely unaffected by the total grid dimensions. Increasing resolution increases memory usage but does not compromise real-time performance. The implementation is fully parallelized using multi-threading and relies exclusively on integer operations.


\begin{figure}[t!]
    \centering
    \includegraphics[width=1.0\linewidth]{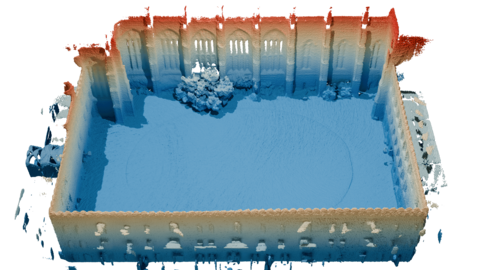}
    \caption{Overview of the reconstructed scene from the Newer College dataset using DB-TSDF.}
    \label{college_tittle}
\end{figure}
   
DB-TSDF builds upon the Truncated Distance Field mapping backend initially developed for the D-LIO framework \cite{DLIO}, which was primarily focused on localization. In this work, the mapping formulation is expanded and refined: the field representation is extended from unsigned to signed distances, directional evidence accumulation is introduced, and the memory layout is optimized for improved cache efficiency. These changes result in a high-resolution mapping method capable of maintaining stable runtimes on CPU-constrained platforms and delivering robust performance even in feature-sparse environments.

The method is evaluated on public LiDAR datasets, with results reported in terms of geometric accuracy, runtime, and the trade-off between resolution, and update latency. The experiments show that DB-TSDF achieves mapping quality comparable to established volumetric approaches while maintaining competitive performance on CPU at resolutions typically requiring GPU acceleration. An example of such high-resolution reconstruction is shown in Figure~\ref{college_tittle}, generated from the Newer College dataset.

The main contributions of this work are: (i) a TSDF integration scheme based on directional kernels and bitmask distance encoding, implemented entirely on CPU; (ii) a mapping approach with constant per-scan computational cost that is independent of the global grid size; (iii) a signed and memory-efficient voxel structure designed for high-resolution mapping and ROS2 integration; and (iv) an experimental evaluation that quantifies the method’s accuracy, speed, and memory requirements.

\section{Related work}

Volumetric mapping methods vary widely in representation, computational requirements, and suitability for real-time deployment. Given the constraints outlined in the introduction, approaches that can operate efficiently on CPU while preserving high resolution are of particular interest. Existing work can be broadly grouped into three categories.

The first group comprises dense, grid-based methods that store explicit volumetric information in fixed-size voxel arrays. These approaches offer deterministic behavior and direct integration into robotic pipelines. Examples include Voxblox \cite{Voxblox}, which produces Euclidean Signed Distance Fields (ESDFs) from Truncated Signed Distance Fields (TSDFs) to support fast planning queries, and FIESTA \cite{FIESTA}, which accelerates occupancy updates through wavefront propagation. Voxfield \cite{Voxfield} further improves reconstruction accuracy with non-projective TSDF fusion and normal estimation from raw point clouds. While such methods are effective on CPU, their dense storage can lead to high memory usage, discretization artifacts, and reduced scalability in large-scale environments.

A second category uses hierarchical or sparse data structures to improve scalability. OpenVDB \cite{OpenVDB} is widely adopted in this space, providing efficient sparse storage and fast random access. Variants such as VDB-EDT \cite{VDB-EDT} and VDBFusion \cite{VDBFusion} integrate distance field computations directly into the VDB framework, while VDBblox \cite{VDBblox} combines the mapping pipeline of Voxblox with OpenVDB's storage efficiency. These methods reduce memory overhead and allow larger maps, but remain limited by the discrete nature of voxel grids, which can introduce aliasing and complicate continuous optimization.

A third group of methods replaces discrete grids with continuous representations. Neural implicit models, such as DeepSDF \cite{DeepSDF} and Neural Unsigned Distance Fields \cite{Neural_unsigned_DF}, encode 3D geometry in neural networks that can be queried at arbitrary resolution. Real-time or incremental variants have been explored in Continual Neural Mapping \cite{Continual_neural_mapping}, iSDF \cite{iSDF}, DI-Fusion \cite{DI-Fussion}, and HIO-SDF \cite{HIO-SDF}. Despite their fidelity, these methods are computationally intensive, often require GPU acceleration, and involve non-trivial integration into robotic systems. Probabilistic models, such as Gaussian Process Implicit Surfaces \cite{Log-GPIS-MOP,GPDF_for_mapping} and VDB-GPDF \cite{VDB-GPDF}, offer uncertainty-aware mapping but face scalability challenges due to the cubic cost of GP inference, even with hierarchical decompositions \cite{DF_mapping_and_planning}.

The original D-LIO framework \cite{DLIO} incorporated a Truncated Distance Field (TDF) mapping backend to support odometry estimation, with the mapping stage designed as a lightweight complement to localization. In contrast, the present work focuses on the mapping problem itself, extending the TDF formulation into a signed representation, introducing direction-dependent kernels. DB-TSDF retains the deterministic and CPU-friendly nature of grid-based methods. For a given input it produces identical reconstructions and captures the full geometric information of each scan from the first frame, while addressing scalability and resolution trade-offs. This makes it a practical solution for high-resolution, real-time volumetric mapping on CPU-only platforms.

\section{FAST TRUNCATED SIGNED DISTANCE FIELD}

The proposed mapping backend builds upon the Fast Truncated Distance Field (Fast-TDF) formulation introduced in D-LIO, originally designed to support odometry estimation with minimal computational overhead. In that formulation, the TDF representation was unsigned, direction-agnostic, and optimized for constant-time CPU updates.

In this work, we extend the concept into a \textit{Directional Bitmask-based Truncated Signed Distance Field} (DB-TSDF), introducing three key enhancements:

\begin{itemize}
    \item \textbf{Signed distance encoding}, enabling explicit differentiation between free and occupied space at the voxel level.
    \item \textbf{Directional kernels}, which model the anisotropic update pattern of LiDAR returns, including a shadow region behind each measured surface point.
    \item \textbf{Compact voxel layout}, minimizing per-voxel memory footprint and optimizing cache locality while preserving constant-time indexing across the entire grid.
\end{itemize}

These modifications shift the focus from odometry support to a dedicated high-resolution volumetric mapping backend. DB-TSDF preserves the deterministic behavior of the original TDF and supports efficient parallelization, producing dense, signed maps in real time on CPU.

\subsection{Voxel representation}

The DB-TSDF map is stored as a dense, axis-aligned voxel grid of fixed resolution. Each voxel encodes three fields:

\begin{itemize}
    \item \textbf{Distance mask}: a 32-bit unsigned integer encoding the truncated $L_1$ distance to the nearest occupied cell within a fixed kernel radius, discretized in voxel cells. The distance is implicitly represented by the number of active bits, enabling constant-time updates via bitwise AND operations, as detailed in Section~\ref{sec:map_update_process}.
    \item \textbf{Sign flag}: a single bit that indicates whether the voxel is classified as occupied ($0$) or free ($1$). This is determined from the directional kernel update, which accounts for the surface hit and the shadow region behind it.
    \item \textbf{Hit counter}: an 8-bit unsigned integer counting the number of observations supporting the occupied state. Once the counter exceeds a predefined threshold, the voxel is considered confirmed as occupied.
\end{itemize}

This compact structure requires only $S_{\mathrm{voxel}} = 8$~bytes per voxel, resulting in a total memory footprint of:

\begin{equation}
    M_{\mathrm{grid}} = N_x \cdot N_y \cdot N_z \cdot S_{\mathrm{voxel}},
    \label{eq:memory_grid}
\end{equation}
where $N_x$, $N_y$, and $N_z$ are the grid dimensions along each axis.  

At initialization, all distance masks are set to their maximum value (all bits set), the sign flag is set to \textit{free}, and hit counters are zeroed. This ensures that kernel applications can only reduce the stored distance and update the occupancy state when supported by new evidence.

The dense memory layout ensures predictable access times and simplifies parallel execution. The regular voxel arrangement allows updates to be distributed across multiple threads without memory contention, enabling high-throughput processing of large point clouds.

\subsection{Directional Kernels}

A key feature of DB-TSDF is the use of \textit{directional kernels} to update the voxel grid according to the orientation of each input point. Unlike the isotropic update in the original D-LIO mapping backend, directional kernels incorporate the anisotropic footprint of the beam and the occlusion pattern it produces.

The directional space around the sensor is discretized into $B_{az} \times B_{el}$ angular bins, with $B_{az} = B_{el} = 40$, resulting in 1600 bins. Each bin corresponds to a representative unit vector pointing at the center of its azimuth–elevation range and is associated with a dedicated precomputed $21 \times 21 \times 21$ cubic kernel. Given a point $\mathbf{p}=(x,y,z)$ expressed in the sensor coordinate frame, the corresponding azimuth and elevation bin indices are computed as:

\begin{equation}
\begin{split}
  b_{\mathrm{az}} &= \left\lfloor \frac{\operatorname{atan2}(y,x)}{2\pi}  B_{\mathrm{az}} \right\rfloor,\\
  b_{\mathrm{el}} &= \left\lfloor \frac{\arcsin\!\left(\frac{z}{\left\| \mathbf{p} \right\|}\right) + \tfrac{\pi}{2}}{\pi}  B_{\mathrm{el}} \right\rfloor.
\end{split}
\label{eq:bin_index}
\end{equation}
where $b_{\mathrm{az}}$ and $b_{\mathrm{el}}$ are the discrete azimuth and elevation indices, $(x,y,z)$ are the Cartesian coordinates of $\mathbf{p}$, $\|\mathbf{p}\|$ is its Euclidean norm, and $B_{\mathrm{az}}, B_{\mathrm{el}}$ are the numbers of bins per angle. Each kernel is aligned with the bin’s representative unit vector, defined by the center of its azimuth–elevation interval.

Within each kernel, every voxel stores a 32-bit mask encoding the truncated $L_1$ distance to the kernel center, discretized in voxel cells:

\begin{equation}
    m(x,y,z) =
    \begin{cases}
    0, & \text{if } r = 0, \\
    (2^{32}-1) \gg (32 - \lceil r \rceil), & \text{if } r > 0,
    \end{cases}
    \label{eq:distance_mask}
\end{equation}
where $r = \sqrt{x^2 + y^2 + z^2}$ is the voxel distance in grid units, and $\gg$ denotes a right bit shift.

\begin{figure*}[t]
    \centering
    \includegraphics[width=0.19\linewidth]{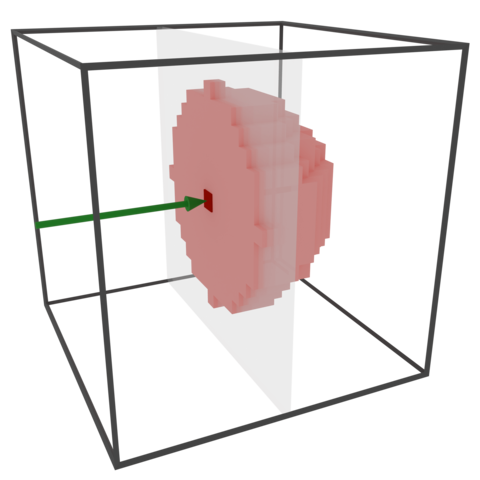}
    \includegraphics[width=0.19\linewidth]{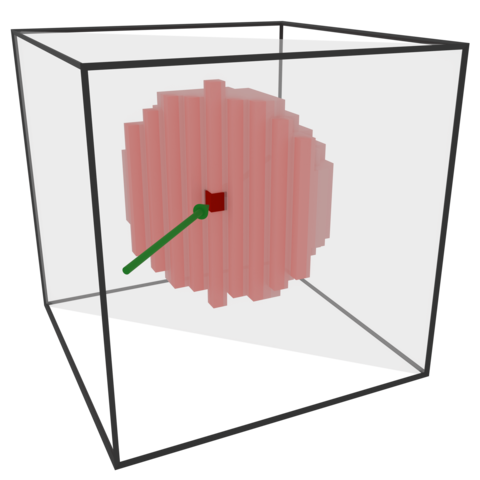}
    \includegraphics[width=0.19\linewidth]{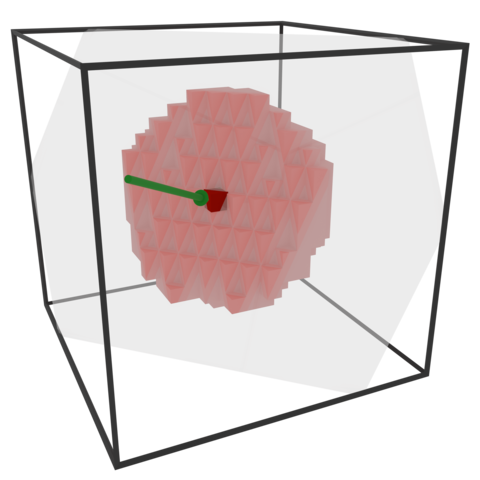}
    \includegraphics[width=0.19\linewidth]{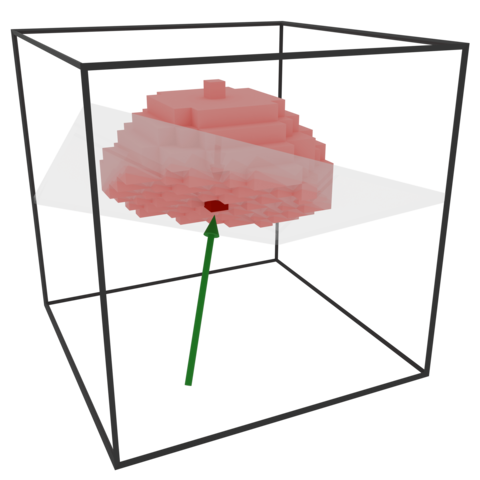}
    \includegraphics[width=0.19\linewidth]{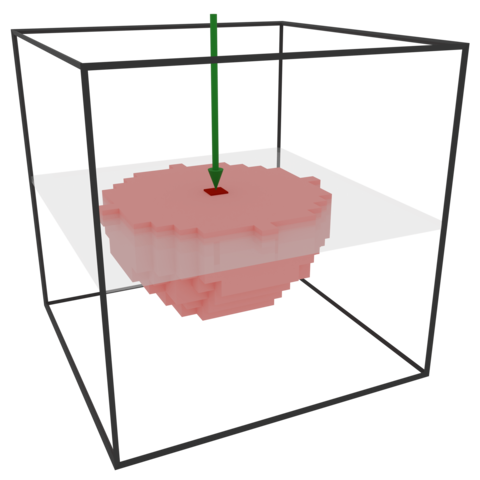}
    \caption{Examples of the hemispherical shadow region (red) inside the $21 \times 21 \times 21$ voxel kernel for different ray directions (green arrows). Each panel shows how the hemispherical mask remains aligned with the incoming LiDAR beam while preserving a consistent exclusion pattern around the contact voxel.}
    \label{directional_kernel}
\end{figure*}

The occupied (shadow) region is modeled as a truncated $L_2$ hemisphere of radius $r_s$ voxels aligned with the bin’s representative direction (Fig.~\ref{directional_kernel}). This corresponds to the set of voxels whose centers lie inside a Euclidean sphere around the contact point, with the front-facing half discarded under the assumption of free space. The $L_2$ model is preferred over an $L_1$ sphere, which reduces to an octahedron and introduces discretization artifacts. The parameter $r_s$ should scale with grid resolution: large values for coarse grids to maintain surface continuity, and smaller ones for fine grids to reduce false positives. Errors at depth discontinuities, such as corners, are naturally corrected by subsequent rays.

\begin{figure}[t!]
    \centering
    \includegraphics[width=0.32\linewidth]{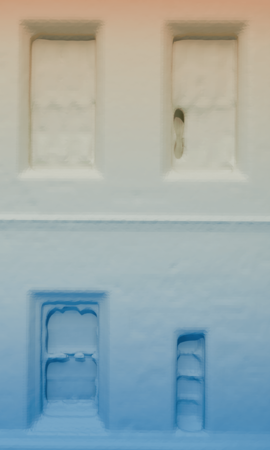}
    \includegraphics[width=0.32\linewidth]{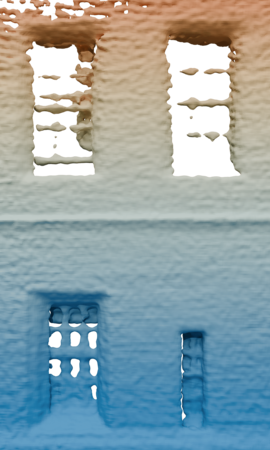}
    \includegraphics[width=0.32\linewidth]{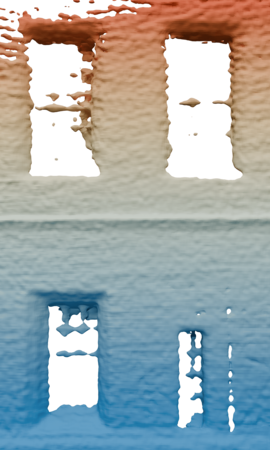}
    \caption{Comparison of shadow models on the Newer College dataset (frames 1–250, voxel size 2.5 cm). Left: ground-truth reconstruction of the Quad at 1 cm resolution. Center: hemispherical model. Right: conical model.}
    \label{fig:cone_vs_sphere}
\end{figure}

In practice, we adopt the hemispherical model because it strengthens occupancy evidence in the early stages of integration, when only a limited number of frames are available. The planar face of the hemisphere, perpendicular to the ray, increases the probability that neighboring voxels of the contact cell also accumulate hits, facilitating earlier confirmation of occupied status. By contrast, a conical shadow restricts hits almost exclusively to the central voxel, making it harder to reach the occupancy threshold unless multiple returns fall directly on the same cell. Consequently, the hemispherical model provides sharper details with fewer frames, while the conical alternative requires more repeated measurements to reach a similar level of surface consistency. For dense point clouds or after many frames, both approaches yield reconstructions of comparable quality. As a reference, we also compare against the 3D reconstruction of the Quad at 1 cm resolution, post-processed from the ground-truth map. This effect is illustrated in Fig.~\ref{fig:cone_vs_sphere}, which compares close-up reconstructions on the Newer College dataset (frames 1–200, voxel size 2.5 cm) using hemispherical and conical kernels, alongside the ground-truth reconstruction.

\subsection{Map Update Process}
\label{sec:map_update_process}

\begin{figure*}[t!]
    \centering
    \includegraphics[width=1.0\linewidth]{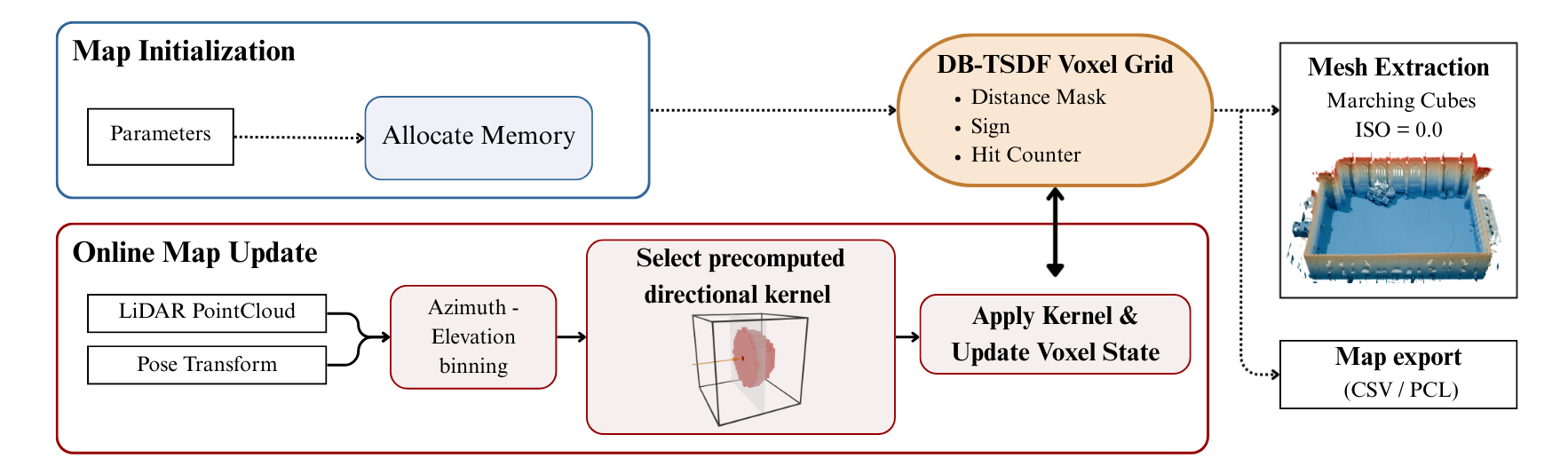}
    \caption{DB-TSDF mapping pipeline. The blue section (Map Initialization) sets parameters and allocates memory for the voxel grid. The red section (Online Map Update) processes each LiDAR return: the point cloud is motion-compensated, quantized to an azimuth–elevation bin, and its corresponding precomputed directional kernel is applied via a binary AND operation with sign and hit-counter updates. The orange block represents the DB-TSDF voxel grid, storing distance mask, sign, and hit counter. The resulting map can be exported as a mesh (Marching Cubes, iso = 0.0) or as CSV/PCL.}
    \label{map_update}
\end{figure*}

The overall DB-TSDF update pipeline is illustrated in Fig.~\ref{map_update}. Each LiDAR return is first transformed into the global map frame using the estimated pose, optionally applying yaw-only or full SE(3) motion compensation. Returns whose $21 \times 21 \times 21$ voxel neighborhood would extend beyond the map boundaries are first discarded. The return’s azimuth–elevation pair is then quantized into one of the $B_{az} \times B_{el}$ bins, selecting the corresponding precomputed directional kernel.

The voxel’s distance mask is updated by a bitwise AND with the selected kernel mask. Because the AND operation is only clears bits, the stored value never increases; the write is skipped if the result is unchanged:

\begin{equation}
m_{\mathrm{grid}} \leftarrow
\begin{cases}
m_{\mathrm{grid}}  \&  m_{\mathrm{kernel}}, & \text{if } m_{\mathrm{grid}} \neq m_{\mathrm{grid}}  \&  m_{\mathrm{kernel}},\\
m_{\mathrm{grid}}, & \text{otherwise.}
\end{cases}
\label{eq:bitwise_and_update}
\end{equation}

Occupancy is determined by a saturating hit counter that is incremented only for voxels in the kernel’s shadow region. Let $h$ denote the counter, $H_{\max}$ the saturation value, and $T$ the occupancy threshold:

\begin{equation}
    h^{\text{new}} \;=\; \min\!\big(h^{\text{old}} + \mathbf{1}_{\text{shadow}},  H_{\max}\big).
\end{equation}

The sign bit $s$ (0=occupied, 1=free) is updated only when the counter crosses the threshold:

\begin{equation}
    s^{\text{new}} \;=\;
    \begin{cases}
      0, & \text{if } h^{\text{old}} < T \text{ and } h^{\text{new}} \ge T,\\[2pt]
      s^{\text{old}}, & \text{otherwise}.
    \end{cases}
    \label{eq:sign_update}
\end{equation}

This removes per-point geometric calculations at runtime, limiting each update to a few integer operations and memory writes. With OpenMP parallelization across points, high update rates are sustained. The per-frame cost is approximated by:

\begin{equation}
    T_{\mathrm{update}} \approx N_p \cdot K^3 \cdot C_{\mathrm{op}},
    \label{eq:update_time}
\end{equation}
where $N_p$ is the number of LiDAR points, $K$ is the kernel size (constant, $K=21$ in our case), and $C_{\mathrm{op}}$ is the average per-operation cost (bitwise AND plus assignments). Since $K$ is fixed and independent of the global map resolution, $T_{\mathrm{update}}$ remains constant when changing the map resolution.

\section{Evaluation}

The DB-TSDF framework is implemented in C++ within the ROS2 middleware and operates entirely on the CPU. All experiments were conducted on a workstation equipped with a 13th Gen Intel\textsuperscript{\textregistered} Core\texttrademark{} i7-13620H processor (16 threads) and 32GB RAM.

Two datasets were used for quantitative evaluation. The Newer College dataset \cite{Newer_college_dataset} provides handheld LiDAR sequences with ground-truth reconstructions obtained from a terrestrial laser scanner. We use the Main Quad sequence (approximately $70 \times 45 \times 30$ m, $\sim$2{,}000 frames). The Mai City dataset \cite{Mai_city} is a synthetic large-scale urban environment of $720 \times 35 \times 7$ m, generated from a 64-beam noise-free LiDAR simulator, providing an idealized reference mesh for evaluation. Both datasets supply ground-truth maps, enabling consistent measurement of accuracy, completeness, and distance errors. For reproducibility, we use the ground-truth assets and ROS 2 bag recordings distributed by SHINE-Mapping to standardize preprocessing across methods \cite{shine_mapping}. Moreover, all quantitative metrics in Tables \ref{tab:mai_city} and \ref{tab:newer_college} are computed with the SHINE-Mapping evaluation pipeline. Metrics for Voxblox, VDB Fusion, PUMA \cite{PUMA}, and SHINE-Mapping are taken directly from the SHINE-Mapping paper \cite{shine_mapping}, whereas the results for VDB-GPDF were obtained by running the available code \cite{VDB_GPDF_repo} in our workstation using the same datasets and evaluation procedure to ensure direct comparability.

\subsection{Mapping results}
\label{subsec:mapping_results}

Following common practice in large-scale LiDAR mapping, we assess reconstruction quality by comparing the predicted point set sampled from zero-level isosurface of the reconstructed mesh against the ground-truth reference. The evaluation is based on nearest-neighbor distances that quantify both surface accuracy and coverage.

Two complementary one-sided errors are computed. Accuracy corresponds to the mean nearest-neighbor distance from each predicted point to the ground truth, penalizing redundant or misplaced geometry. Completeness is defined symmetrically as the mean nearest-neighbor distance from each ground-truth point to the prediction, penalizing missing regions of the reconstruction. Both values are expressed in meters, and lower values indicate better performance. Chamfer L1 is the average of Accuracy and Completeness, providing a single symmetric distance measure that balances geometric precision and surface coverage.

In addition, we report thresholded measures that quantify reconstruction fidelity at a tolerance $t$. Recall indicates the proportion of ground-truth points that are successfully recovered by the reconstruction within the tolerance, and therefore reflects how completely the reference surface is covered. The F-score combines coverage and precision into a single balanced score, providing an interpretable measure of overall reconstruction quality. A high Recall implies minimal missing regions, while a high F-score denotes a reconstruction that is both complete and consistent with the ground truth.

\begin{table}[t]
\caption{Quantitative results on \textbf{Mai City}. Acc.\ Comp.\ and C-L1 in cm (lower is better); Recall and F-score at 10 cm (higher is better). The best and second-best results in each column are shown in \textbf{bold}, and \underline{underlined}, respectively.}
\label{tab:mai_city}
\centering
\begin{tabular}{@{}lccccc@{}} 
\toprule
Method & Acc. $\downarrow$ & Comp. $\downarrow$ & C-L1 $\downarrow$ & Recall $\uparrow$ & F-score $\uparrow$ \\
\midrule
VDB-GPDF         & 3.8             & 8.5             & 6.2             & 83.3             & 88.7 \\
Voxblox          & 1.8             & 7.1             & 4.8             & 84.0             & 90.9 \\
VDB Fusion       & 1.3             & 6.9             & 4.5             & 90.2             & 94.1 \\
PUMA             & \underline{1.2} & 32.0            & 16.9            & 78.8             & 87.3 \\
SHINE-Mapping    & \textbf{1.1}    & \textbf{3.2}    & \textbf{2.9}    & \textbf{95.2}    & \underline{95.9} \\
\midrule
DB-TSDF (ours)   & 1.7             & \underline{4.6} & \underline{3.1} & \underline{93.6} & \textbf{96.6} \\
\bottomrule
\end{tabular}
\end{table}

\begin{figure}[t!]
    \centering
    \includegraphics[width=\linewidth]{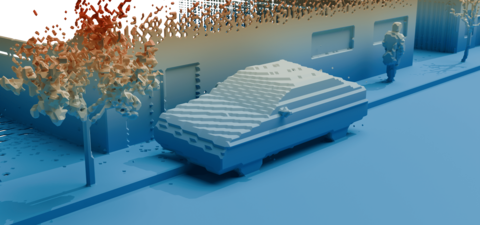}\\[4pt]
    \includegraphics[width=\linewidth]{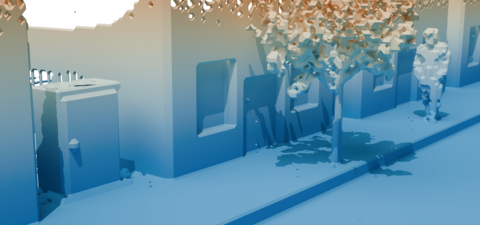}
    \caption{Close-up reconstructions from the Mai City dataset using DB-TSDF, illustrating surface fidelity and continuity in different regions of the scene.}
    \label{fig:mai_city_detail}
\end{figure}

\begin{table}[t]
\caption{Quantitative results on \textbf{Newer College}. Acc.\ Comp.\ and C-L1 in cm (lower is better); Recall and F-score at 20 cm (higher is better). The best and second-best results in each column are shown in \textbf{bold}, and \underline{underlined}, respectively.}
\label{tab:newer_college}
\centering
\begin{tabular}{@{}lccccc@{}} 
\toprule
Method & Acc. $\downarrow$ & Comp. $\downarrow$ & C-L1 $\downarrow$ & Recall $\uparrow$ & F-score $\uparrow$ \\
\midrule
VDB-GPDF       & 7.5             & 11.9             & 9.7             & 91.7             & 90.4 \\
Voxblox        & 9.3             & 14.9             & 12.1            & 87.8             & 87.9 \\
VDB Fusion     & \underline{6.9} & 12.0             & \underline{9.4} & 91.3             & \underline{92.6} \\
PUMA           & 7.7             & 15.4             & 11.5            & 89.9             & 91.9 \\
SHINE-Mapping  & \textbf{6.7}    & \textbf{10.0}    & \textbf{8.4}    & \textbf{93.6}    & \textbf{93.7} \\
\midrule
DB-TSDF (ours) & 9.1             & \underline{10.7} & 9.9             & \underline{92.4} & 91.3 \\
\bottomrule
\end{tabular}
\end{table}

On the Mai City dataset (Table~\ref{tab:mai_city}), DB-TSDF achieves state-of-the-art performance with the highest F-score of 96.6\%. It further ranks second in completeness (4.6 cm), Chamfer L1 (3.1 cm), and recall (93.6\%), showing both geometric accuracy and surface coverage competitive with the best available methods. In particular, edges and corners of the environment are clearly preserved, reflecting the ability of the directional kernels to model occlusion patterns and the bitmask encoding to maintain sharp geometric structures without introducing artifacts.

On the Newer College dataset (Table~\ref{tab:newer_college}), DB-TSDF attains the second-best results in completeness (10.7 cm) and recall (92.4\%), and ranks third in Chamfer L1 (9.9 cm), confirming its robustness in real-world conditions with sensor noise and environmental variability. Additional qualitative results and visualizations of the reconstructed TSDF meshes are provided in the supplementary material.

\subsection{Efficiency}

\begin{figure}[t]
    \centering
    \includegraphics[width=\linewidth]{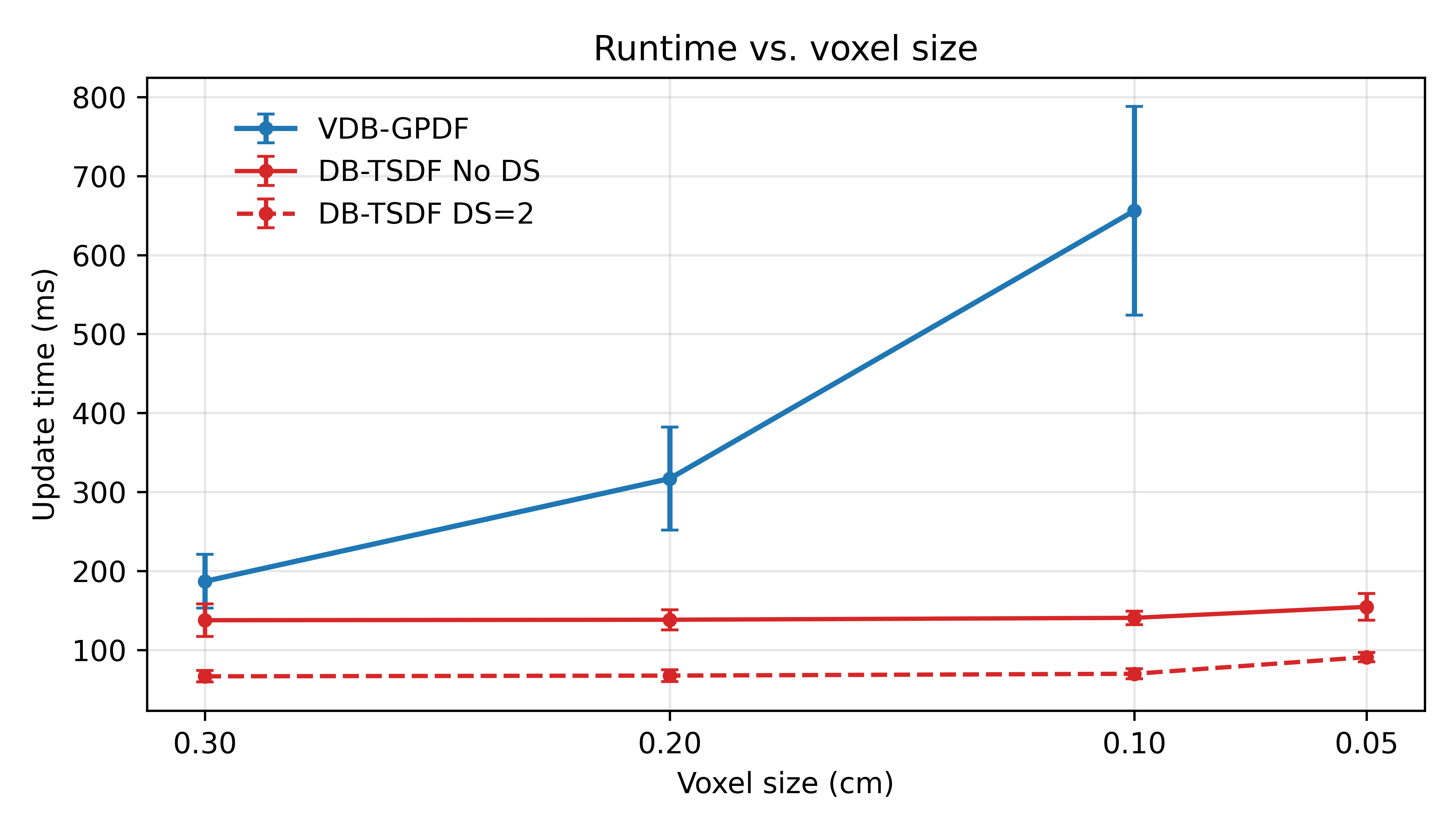}
    \caption{Average runtime per LiDAR frame versus voxel size for
    VDB\textendash GPDF, DB\textendash TSDF without downsampling (No DS), 
    and DB\textendash TSDF with DS=2.}    \label{fig:efficiency_plot}
    \label{tab:time_metrics}
\end{figure}

The computational performance of the proposed method was assessed on the Newer College dataset. The average runtime per LiDAR frame remains stable at approximately 150 ms across voxel resolutions ranging from $0.3$ m to $0.05$ m (Fig.\ref{fig:efficiency_plot}), with $154.8\pm16.9$ ms observed at $0.05$ m without downsampling. This near resolution-invariance stems from the bitmask-based update; additionally, for coarser voxels than the point-cloud spacing, several returns fall into the same voxel within a frame and only the first is integrated, slightly reducing latency. Consequently, the number of operations per LiDAR return is bounded and independent of the global grid resolution. The computational cost is therefore dominated by the number of points per scan, while changes in voxel size only affect memory usage. 

\begin{figure}[t!]
\centering
\includegraphics[width=\linewidth]{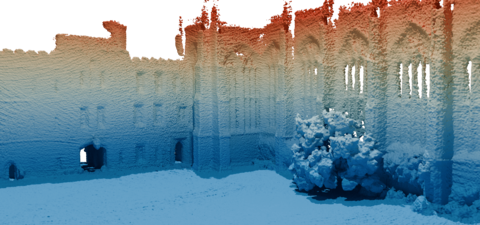}\\ [4pt]
\includegraphics[width=\linewidth]{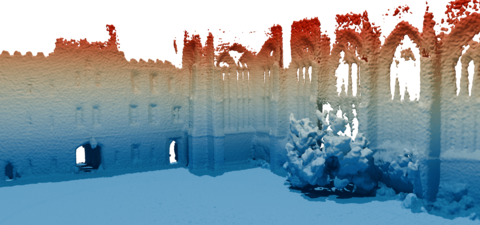}
\caption{Newer College meshes at $0.05$ m. Top: no downsampling (No DS); bottom: DS$=2$.}
\label{fig:college_mesh_ds_compare}
\end{figure}

Real-time operation can be achieved with light sequential downsampling. At $0.05$ m, DS=2 reduces per-frame latency to $91.1\pm$ ms. The reconstruction remains broadly consistent with the no-downsampling baseline; for DS=2 (20 cm tolerance) we obtain Acc./Comp./C-L1 of $11.4/13.1/12.3$ cm and Recall/F-score of $92.7\%/87.4\%$ (Fig.~\ref{fig:college_mesh_ds_compare}). 

For context, CPU-based probabilistic volumetric methods such as VDB-GPDF~\cite{VDB-GPDF} report runtimes above 500 ms per LiDAR frame at a voxel resolution of 0.1 m. These values were measured on our workstation to ensure a fair CPU-to-CPU comparison (see Fig~\ref{tab:time_metrics}). In contrast, DB-TSDF achieves an average of 150 ms while delivering competitive geometric accuracy. Although SHINE-Mapping achieves top scores in some accuracy metrics, it relies on GPU acceleration and is therefore not included in the runtime comparison. This performance gain is obtained with a simpler and fully deterministic bitmask-based formulation. Unlike learning-based approaches, the proposed integration provides an accurate environment representation from the very first scan, without requiring training or prior data.

\section{CONCLUSIONS}

High-resolution volumetric mapping on CPU is often constrained by computational cost and scalability issues. DB-TSDF shows that dense reconstructions can still be achieved efficiently without relying on GPU acceleration, through a compact and deterministic design optimized for fast updates.

By combining bitmask-based distance encoding with precomputed directional kernels, the method achieves constant per-frame update times, independent of the map size. This allows high-resolution maps to be maintained using only integer operations and multi-threaded parallelization.

Qualitative evaluation confirms that DB-TSDF delivers mapping quality on par with established volumetric methods across synthetic and real-world datasets, remaining robust even under sensor noise and environmental variability. At the same time, efficiency is preserved, with runtimes that remain constant and reach at most around 150 ms per LiDAR frame, or lower when downsampling is applied, making the approach well suited for real-time deployment on CPU-only robotic platforms where GPU resources are limited or dedicated to other tasks.

Future work will extend the framework to dynamic environments and explore adaptive kernels for improved reconstruction at depth discontinuities.

\addtolength{\textheight}{-12cm}   



\balance
\bibliographystyle{IEEEtran}
\bibliography{bibliography}

\end{document}